%% file: sample-sigconf.tex
\documentclass[sigconf]{acmart}
\AtBeginDocument{%
  }
\setcopyright{acmcopyright}
\copyrightyear{2023}
\acmYear{2023}
\acmDOI{XXXXXXX.XXXXXXX}

\acmConference[Multimodal KDD '23]{29TH ACM SIGKDD CONFERENCE ON KNOWLEDGE DISCOVERY AND DATA MINING}{August 07, 2023}{Long Beach, CA}
\acmBooktitle{29TH ACM SIGKDD CONFERENCE ON KNOWLEDGE DISCOVERY AND DATA MINING (Multimodal KDD '23), August 07, 2023, Long Beach, CA}
\acmPrice{15.00}
\acmISBN{978-1-4503-XXXX-X/18/06}

\input{math_commands}

\begin{document}

\title{What Makes Good Open-Vocabulary Detector: A Disassembling Perspective}

\author{Jincheng Li}
\authornote{Both authors contributed equally to this research.}
\affiliation{%
  \institution{Qihoo 360 AI Research}
  \city{Beijing}
  \country{China}}
\email{lijincheng@360.cn}

\author{Chunyu Xie}
\authornotemark[1]
\affiliation{%
  \institution{Qihoo 360 AI Research}
  \city{Beijing}
  \country{China}}
\email{xiechunyu@360.cn}

\author{Xiaoyu Wu}
\affiliation{%
  \institution{Qihoo 360 AI Research}
  \city{Beijing}
  \country{China}}
\email{wuxiaoyu1@360.cn}

\author{Bin Wang}
\affiliation{%
  \institution{Qihoo 360 AI Research}
  \city{Beijing}
  \country{China}}
\email{wangbin10@360.cn}

\author{Dawei Leng}
\authornote{Corresponding Author.}
\affiliation{%
  \institution{Qihoo 360 AI Research}
  \city{Beijing}
  \country{China}}
\email{lengdawei@360.cn}


\begin{abstract}
    Open-vocabulary detection (OVD) is a new object detection paradigm, aiming to localize and recognize unseen objects defined by an unbounded vocabulary. This is challenging since traditional detectors can only learn from pre-defined categories and thus fail to detect and localize objects out of pre-defined vocabulary. To handle the challenge, OVD leverages pre-trained cross-modal VLM, such as CLIP, ALIGN, etc. Previous works mainly focus on the open vocabulary classification part, with less attention on the localization part. We argue that for a good OVD detector, both classification and localization should be parallelly studied for the novel object categories. We show in this work that improving localization as well as cross-modal classification complement each other, and compose a good OVD detector jointly. We analyze three families of OVD methods with different design emphases. We first propose a vanilla method, \ie cropping a bounding box obtained by a localizer and resizing it into the CLIP. This vanilla method totally decouples the localization and classification components, making it convenient to improve the OVD performance by applying more advanced object localization models and VLMs. However, resizing cropped regions inevitably causes the deformation of the object and leads slow calculation speed. To address these, we next introduce another approach, which combines a standard two-stage object detector with CLIP. A two-stage object detector includes a visual backbone, a region proposal network (RPN), and a region of interest (RoI) head. We decouple RPN and ROI head (DRR) and use RoIAlign to extract meaningful features. In this case, it avoids resizing objects. To further accelerate the training time and reduce the model parameters, we couple RPN and ROI head (CRR) as the third approach. We conduct extensive experiments on these three types of approaches in different settings. On the OVD-COCO benchmark, DRR obtains the best performance and achieves 35.8 Novel AP$_{50}$, an absolute 2.8 gain over the previous state-of-the-art (SOTA). For OVD-LVIS, DRR surpasses the previous SOTA by 1.9 AP$_{50}$ in rare categories. We also provide an object detection dataset called PID and provide a baseline on PID. 
\end{abstract}

\ccsdesc[500]{Computing methodologies~Object detection}


\keywords{vision-language model, open-vocabulary detection, dataset}


\maketitle

\section{Introduction}

\ljc{
Object detection is a prominent vision task, aiming at localizing and recognizing objects in images. This task requires a variety of fine-grained annotations (\eg the bounding boxes and classes) of each object during training, which, however, makes it hard to extend the size of data since manual human annotations are costly and tedious. In this sense, traditional object detectors may fail to precisely detect and localize objects out of pre-defined vocabulary at inference.
}

\ljc{
Open-vocabulary detection (OVD), a task to detect unseen objects defined by an unbounded vocabulary, has attracted much attention in the most recent period. The core challenge of the OVD task is how to localize and classify unseen (novel) categories at the inference stage since they can only learn the knowledge from pre-defined (base) categories during training.
We next analyze the corresponding existing solutions to overcome the above challenge from two perspectives: classification and localization.
}

\xie{
To classify the novel categories, several works \cite{ViLD, regionclip, bangalath2022bridging} leverage the excellent zero-shot generalization ability of large-scale vision-language models (VLMs) such as CLIP \cite{CLIP}. To this end, they modify a two-stage object detector, including using the visual encoder of CLIP as the backbone of the object detector, replacing the class-specific classification head with a class-agnostic classification head, and so on.
Meanwhile, they re-train or finetune VLMs for open-vocabulary detection since most VLMs are pre-trained on image-text pairs but not detection data. While detection-tailored pre-training is beneficial for the OVD task, some studies like F-VLM \cite{kuo2022f} discard this technique and also achieve significant results.
In this sense, whether to use detection-tailored pre-trained CLIP remains an open question to be discussed.
}

\xie{
On the other hand, to localize the novel categories, most existing methods \cite{regionclip, ViLD, BARON} use a region proposal network (RPN), which is trained on base categories. They demonstrate that the RPN trained without seeing novel categories can generalize to localize novel categories \cite{ViLD}. The bounding boxes obtained by RPN play a critical role in predicting the final boxes. In this way, a better proposal network should further improve the overall performance of the OVD model since the generated bounding boxes will be more accurate.
However, how to effectively improve the detection ability under the settings of the OVD task is still a challenge to be solved.
}

\ljc{
In this paper, we set out to address these issues under the settings of the open-vocabulary detection task. Our goal is to analyze which part of localization and classification can improve the overall performance of the OVD task.
We use three simple but effective approaches, discuss their advantages and disadvantages and design appropriate experiments for them.
}

\ljc{
We first introduce a vanilla method for the OVD task, \ie cropping the bounding box, resizing it into the required input size of CLIP, and then feeding it into the visual encoder of CLIP. The detected bounding box is given by a pre-trained RPN, while the classification score is calculated by the cosine similarity of box embedding and the text embedding, which is extracted by the text encoder of CLIP. This vanilla method completely decouples the detection and classification components, making it easier to \xie{adopt} different models.
In this case, we apply different RPNs to investigate the effectiveness of RPN under the OVD settings, consisting of RPN with extra data or stronger RPN. Our goal in this experiment is to improve the detection ability, such that the overall performance of the OVD task would be further improved.
In addition, we attempt to use different types of CLIP to enhance the vanilla performance.
Although it is convenient to replace each component, it still suffers from two limitations: 1) it is non-trivial to extract meaningful features when facing tiny bounding boxes, \ie $2 \times 2$ pixel region. 2) it has slow inference speed due to the operations of cropping and resizing.
}

\ljc{
Next, we present another popular approach, which modifies a standard two-stage object detector \cite{fasterrcnn} and combines it with CLIP.  A two-stage object detector consists of a visual backbone, an RPN, and a region of interest (RoI) \xie{head}. Several works \cite{ViLD,kuo2022f,BARON} use a pre-trained RPN to extract proposals, then obtain their features via the visual encoder of CLIP relying on RoIAlign \cite{fasterrcnn}.
\xie{The characteristic of this approach is that it decouples RPN and ROI head (DRR) to reduce the fusion of detection features and classification features.}
Compared to the vanilla method, \xie{\xie{DRR} is able to extract significant features for tiny objects and has a faster inference speed.} Unfortunately, it applies two backbones to decouple RPN and \xie{ROI head}, leading to require more computational costs and training time.
}

\ljc{
\xie{Based on DRR, we propose the third approach, which couples RPN and ROI head (CRR). CRR} uses one backbone for localization and classification. In this sense, it accelerates the training time and reduces the parameters of the model. To summarize, our main contributions are as follows:
}

\begin{itemize}[leftmargin=*]
    \item \ljc{
    We investigate and analyze the advantages and disadvantages of three approaches for open-vocabulary detection. We demonstrate that both localization and classification can improve open-vocabulary detectors.
    }
    \item \ljc{We design appropriate experiments for three approaches with the commonly used techniques and achieve state-of-the-art results on \xie{OVD-COCO and OVD-LVIS} benchmarks.}
    \item \ljc{We propose a product dataset (PID) with human annotations for the open-vocabulary detection task and provide a strong baseline on PID.}
\end{itemize}

\begin{figure*}[t]
    \centering
    \includegraphics[width=0.99\textwidth]{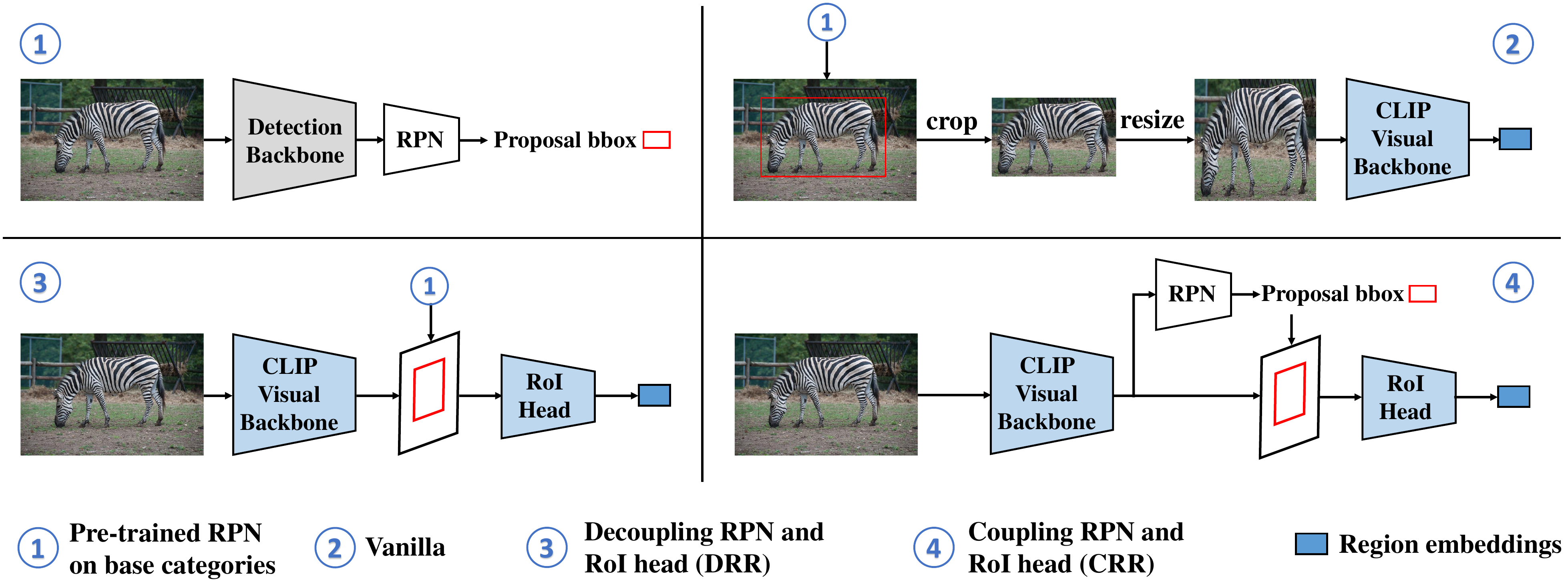}
    \caption{\ljc{An overview of three approaches of open-vocabulary detection: a vanilla method, decoupling RPN and ROI head (\xie{DRR}), and coupling RPN and ROI head (\xie{CRR}). Note that Vanilla and \xie{DRR} require two backbones while \xie{CRR} only needs one visual backbone. For simplicity, we omit the logits of regions when training RPN on base categories. For all three fundamental approaches, we get the classification scores via cosine similarity, calculated by the region embeddings and the text embeddings, where the text embeddings are obtained by the text encoder of CLIP.}} 
    \label{fig:overview}
\end{figure*}

\section{Related Work}

\ljc{
\textbf{Vision-Language Pre-training.}
Vision-language pre-training aims to learn the correspondence between vision and natural language. 
It is attractive that pre-trained vision-language models (VLMs) \cite{CLIP, jia2021scaling, yang2022chinese, xie2022zero} trained on large-scale image-text pairs show excellent zero-shot/few-shot migration ability on classification, object detection, and instance segmentation tasks. In particular, F-VLM \cite{kuo2022f} shows that pre-trained VLMs have a strong generalization ability when transferring to the OVD task.
Meanwhile, F-VLM eliminates the need for knowledge distillation or detection-tailored pre-training. This motivates us to investigate how to make better use of VLMs on the open-vocabulary object task. In this paper, we apply different VLMs in the fundamental approaches, including original VLMs, larger VLMs, and detection-tailored pre-training VLMs.
}





\ljc{
\textbf{Open-Vocabulary Detection.}
Traditional object detection may fail to localize and recognize unseen objects in an image at inference, \ie zero-shot object detection.
Recently, OVR-RCNN \cite{ovrcnn} proposes the open-vocabulary detection (OVD) benchmark to detect and localize objects for which no bounding box annotation is provided during training.
While OVR-RCNN evaluates the models on tens of categories, ViLD \cite{ViLD} proposes to evaluate on more than 1,000 categories, \ie LVIS \cite{LVIS}.
Following the OVD benchmark, most existing methods are proposed with different forms of weak supervision, such as extra image-caption pairs \cite{ovrcnn, regionclip, gao2022open, wu2023cora}, extra classification datasets \cite{Detic}, and vision-language pre-trained models \cite{ViLD, feng2022promptdet, chen2022open, zhao2022exploiting} like CLIP \cite{CLIP}.
For example, RegionCLIP \cite{regionclip} introduces a region-level pre-training alignment method with extra image-text pairs \eg CC3M \cite{CC3M}, demonstrating its capability on zero-shot and OVD task transfer learning. Detic \cite{Detic} trains the classifiers of a detector on image classification data, \ie ImageNet-21K \cite{imagenet}, yielding excellent detectors even for classes without box annotations. BARON\cite{BARON} proposes to align the embedding of the bag of regions beyond individual regions, relying on the generalization ability of large-scale vision-language pre-trained models. These methods keep achieving better results than previous state-of-the-art methods with various techniques, such as more powerful offline proposal generators \cite{wang2023detecting}, knowledge distillation \cite{ma2022open}, and prompt learning \cite{du2022learning}. When facing a new dataset for new scenarios, it leaves researchers and engineers confused about which approaches or techniques to use. In this paper, we summarize three fundamental approaches for open-vocabulary detection and investigate them with different techniques, showing surprising results of different combinations.
}

\section{Approach}
\label{approach}

\ljc{
In this paper, we introduce three approaches for open-vocabulary detection: a vanilla method, decoupling RPN and ROI head (\xie{DRR}), and coupling RPN and ROI head (\xie{CRR}). The overview of the approaches is illustrated in Figure \ref{fig:overview}.
}

\subsection{A Vanilla Method}

\wxy{To solve the open-vocabulary detection problem, we attempt to isolate the open-vocabulary detection task into two independent sub-tasks, that is, object localization and object classification.}

\textbf{Object Localization.}
\ljc{To localize objects,} both \ljc{the} class-aware object detector, \ljc{such as YOLO \cite{redmon2016you}, Faster R-CNN }, and the class-agnostic localizer, such as RPN, OLN\cite{OLN} can be used. 
The first challenge for OVD task is to localize novel objects. To solve this problem, \ljc{ we adopt the detection backbone with RPN to localize all objects and classify them as the foreground class. Furthermore, we improve RPN using Faster R-CNN. Here, the multi-classes head within Faster R-CNN is changed to a class-agnostic head so that it can resolve a binary task like RPN. The class-agnostic module within RPN or modified Faster R-CNN can generalize to novel objects \cite{ViLD}. Besides, we can use a classification-free network such as OLN \cite{OLN} for the purpose of localizing objects. }
\xie{In this case, we can modify the training loss to estimate the objectness of each region purely since localization-related metric tends to be robust to novel objects in the open world~\cite{OLN}.}


\ljc{
\textbf{Object Classification.} After localizing object candidates, we leverage a pre-trained large-scale vision-language model (VLM) such as CLIP to classify them. Specifically,} We crop and resize the \ljc{object} candidates, and feed them into the visual encoder of VLM to achieve \ljc{the corresponding region embeddings.} In order to provide more context cues, these region embeddings are ensemble from \xie{1× crop (crop the image according to the bounding box) and 1.5× crop (extend the crop size to 1.5 times)}. We feed the category texts with a set of prompt templates and then feed them into the text encoder of VLM to obtain the text embeddings. 
Finally, The cosine similarity is calculated by the \xie{averaged text embeddings and region embeddings}, and the per-class NMS is adopted to \xie{obtain the final detection results.}



\xie{As this solution decomposes the OVD task into the object localization and classification sub-tasks completely, it can be easily assembled by different detectors and VLMs, leading to a powerful performance. In the vanilla method, each cropped object region needs to be fixed to the same scale for the VLMs. On the one hand, these operations reduce the efficiency of the model. On the other hand, the deformation of the regions brings more difficulties to object detection, especially for small targets.}

\subsection{Decoupling RPN and ROI head (\xie{DRR})}\label{roialign}

\ljc{
A two-stage object detector consists of a visual encoder backbone, a region proposal network (RPN), and a region of interest (RoI) \xie{head}. Here, the RPN is trained end-to-end to generate high-quality region proposals and then provided for detection in the ROI head.
}
\ljc{
While the traditional two-stage object detector jointly trains RPN and the ROI head, several works \cite{regionclip, wang2023detecting} propose to decouple them to avoid the conflict that the sensitivity of the classification head to novel categories hampers the universality ability of the RPN head. So far, there are no detailed discussions about the effect of whether to decouple RPN and ROI head or not in existing works. In this paper, we try to analyze them and hope to provide a view of the balance of model performance and cost for researchers.
}

\ljc{
As aforementioned, we first introduce a training approach that decouples the RPN and ROI head in a two-stage object detector. Following existing works \cite{regionclip, wang2023detecting, bangalath2022bridging}, we use different backbones for \xie{localization and classification}. That is, one backbone is designed for RPN and the other is for ROI head. Given an image, we first use a pre-trained RPN to obtain a set of regions of interest (\ie proposals). We then use the visual encoder (\eg ResNet-50) of CLIP \cite{CLIP} to encode the whole image. Given proposal boxes of an image, we extract their features along the first 3 blocks based on the whole image encoding and pool them using RoIAlign. The pooled features are then encoded by the last block of the visual encoder. We also replace the class head of Fast R-CNN with the text embeddings encoded by the CLIP text encoder. Considering the setup of open-vocabulary detection \cite{ovrcnn}, we use the base class and hand-crafted prompt as the input of the text encoder of CLIP during training. During inference, we replace the base class with a combination of base and novel classes to generate new text embeddings.
}

\subsection{Coupling RPN and ROI head (\xie{CRR})}
\label{two-way-distillation}
Decoupling proposal generation and ROI head is an efficacious method to keep the universality ability of the proposal generation stage since the proposal generation stage has a class-agnostic classification, which can be easily extended to novel classes. However, this decoupling scheme means that RPN and ROI modules use different backbone encoders, which increases the training cost. In the stage of model deployment and inference, this solution will also bring a lot of extra computing time. Considering the computational efficiency, we practice the scheme of sharing the visual backbone and conduct detailed experiments. These experiments can provide researchers with speed-accuracy trade-offs.

\begin{table*}[t]
\caption{\ljc{Performance on OVD-COCO compared with state-of-the-art methods.}}
      \centering
      \label{sota-coco}
      \resizebox{0.7\textwidth}{!}{
        \begin{tabular}{lcccccc}
        \toprule
        Method & Extra Dataset & Backbone  & Novel AP$_{50}$ & \lightgray{Base AP$_{50}$} & \lightgray{Overall AP$_{50}$}  \\
        \midrule
        OVR-CNN \cite{ovrcnn} & COCO Captions & ResNet50  & 22.8 & \lightgray{46.0} & \lightgray{39.9} \\
        ViLD \cite{ViLD} & -  & ResNet50 & 27.6 & \lightgray{59.5} & \lightgray{51.2} \\
        Detic \cite{Detic} & COCO Captions & ResNet50 & 27.8 & \lightgray{47.1} & \lightgray{45.0} \\
        RegionCLIP \cite{regionclip} & CC3M & ResNet50 & 31.4 & \lightgray{57.1} & \lightgray{50.4} \\
        BARON \cite{BARON} & COCO Captions & ResNet50 & 33.1 & \lightgray{54.8} & \lightgray{49.1} \\
        \midrule
        Vanilla (Ours) & - & ResNet50 &   31.8 & \lightgray{37.2} & \lightgray{35.5} \\
        \xie{CRR} (Ours) & CC3M & ResNet50 & 32.0 & \lightgray{52.5} & \lightgray{47.1} \\
        \xie{DRR} (Ours) & CC3M & ResNet50 & \textbf{35.8}  & \lightgray{54.6} & \lightgray{49.6} \\
        \bottomrule
        \end{tabular}
        }
\end{table*}

\begin{table*}[t]
\caption{\ljc{Performance on OVD-LVIS compared with  state-of-the-art methods.}}
      \centering
      \label{sota-lvis}
      \resizebox{0.7\textwidth}{!}{
        \begin{tabular}{lcccccc}
        \toprule
        Method & Backbone & Require Novel Class \cite{BARON} & AP$_r$ & \lightgray{AP$_c$} & \lightgray{AP$_f$} & \lightgray{mAP}  \\
        \midrule
        RegionCLIP \cite{regionclip} & ResNet50 & $\no$ & 17.1 & \lightgray{27.4} & \lightgray{34.0} & \lightgray{28.2} \\
        ViLD \cite{ViLD} & ResNet50 & $\yes$ & 16.7 & \lightgray{26.5} & \lightgray{34.2} & \lightgray{27.8} \\
        BARON \cite{BARON} & ResNet50 & $\yes$ & 20.1 & \lightgray{28.4} & \lightgray{32.2} & \lightgray{28.4} \\
        \midrule
        Vanilla (Ours) & ResNet50 & $\no$ & 17.2 & \lightgray{14.8} & \lightgray{11.5} & \lightgray{13.9}\\
        \xie{CRR} (Ours) & ResNet50 & $\no$ & 14.0 & \lightgray{23.7} & \lightgray{28.5} & \lightgray{21.9} \\
        \xie{DRR} (Ours) & ResNet50 & $\no$ & 20.1 & \lightgray{29.9} & \lightgray{35.7} & \lightgray{30.5}  \\
        \xie{DRR} (Ours) & ResNet50 & $\yes$ & \textbf{22.0} & \lightgray{25.4} & \lightgray{33.7} & \lightgray{28.1}  \\
        \bottomrule
        \end{tabular}
        }
\end{table*}

\section{Experiments}

\textbf{Dataset and Metrics.} 
\ljc{We comprehensively evaluate three fundamental approaches of OVD task on COCO \cite{COCO}, LVIS \cite{LVIS}, and our proposed Product Image Dataset (PID) benchmarks.}
COCO is a standard dataset comprising 80 categories of common objects in a natural context. It contains 118k images with bounding boxes and instance segmentation annotations. We follow \ljc{OVR-CNN \cite{ovrcnn}} to split the object categories into 48 base categories and 17 novel categories. We also follow \ljc{ViLD \cite{ViLD}} for the LVIS dataset to split the 337 rare categories into novel categories and the rest common and frequent categories into base categories. For simplicity, we denote the open-vocabulary benchmarks based on COCO and LVIS as OVD-COCO and OVD-LVIS\ljc{, respectively}. \ljc{Following ViLD \cite{ViLD}, the Novel AP$_{50}$ and AP$_r$ are the main metric on OVD-COCO and OVD-LVIS, respectively.}
Our open-vocabulary object detectors are trained on base classes. \ljc{Besides}, we split PID into the base and novel categories. The detailed introductions can refer to Sec~\ref{PID}.

The dataset used for detector pre-training is established from the BigDetection dataset~\cite{bigdetection}. As the BigDetection dataset has nearly covered all categories in COCO, we remove the COCO images and delete both novel and base categories of the COCO dataset from the annotations. Some categories with less than 100 train samples are removed as well. We also exclude LVIS data from the BigDetection dataset. Finally, we establish this new BigDetection dataset named BigDetection* (BD*) which has 489 categories for detector pre-training.

\textbf{Implementation Details of the vanilla method.}
\xie{In the OVD-COCO and OVD-LVIS experiments, We use Faster R-CNN as the object detector and frozen CLIP as the object classifier. SGD optimizer is adopted for the 8× training schedule. We train the detector for 720k iterations and divide the learning rate by 10 at 660k and 700k iterations. We also adopt linear warmup for the first 1,000 iterations starting from a learning rate of 0 to 0.02.
Then, we fine-tune the model on base categories of OVD-COCO and OVD-LVIS separately. For OVD-COCO, we train 90k iterations and scale down the learning rate at 60k and 80k. For OVD-LVIS, we train 180k iterations and scale down at 120k and 160k. 
For these two datasets, the learning rate increases to 0.0002 for the first 5k iterations for the warmup.} 

\xie{In the PID experiments, we use Faster R-CNN pre-trained on the BigDetection dataset \cite{bigdetection} as the object detector. ProductCLIP with prompt tuning, which is introduced in Sec \ref{PID_Ablation}, is applied as the object classifier. The detectors are fine-tuned with the 1x training schedule. A warm-up step with a learning rate of 0.001 is performed for the first 400 iterations. 
On both the pre-training and fine-tuning stages, we select 16 samples per GPU with the class-aware sampler. Multi-scale training is adopted with the short edge in the range [640, 800] and the long edge up to 1333. We use 8 A100 GPUS to perform the experiments. }


\textbf{\xie{Implementation Details of \xie{DRR} and \xie{CRR}.}}
\ljc{In Figure \ref{fig:overview}, the detection backbone and the CLIP visual backbone are both ResNet50. We adopt ImageNet \cite{imagenet} and RegionCLIP \cite{regionclip} pre-trained parameters for the detection backbone and the CLIP visual backbone, respectively. During training, we first train the RPN to obtain the proposal bounding boxes on the base categories. We then use a Faster R-CNN  with ResNet50-C4 architecture as the detector and 1x training schedule (90k iterations). We train the detector using the offline pre-trained RPN. That is, we decouple the RPN and the ROI head. We use an SGD optimizer with a learning rate of 0.002 and a min-batch of 16.
We set the weight of the background category to 0.2 and 0.8 for OVD-COCO and OVD-LVIS, respectively.
We use the same learning rate and input-scale strategy as the vanilla method.
We replace the base categories in the classification head with base + novel categories during inference. We use the top-ranked 100 proposals at test time for all detectors.}
\ljc{For \xie{CRR}, we keep the same settings as \xie{DRR}, except that \xie{CRR} removes the detection backbone, as shown in Figure \ref{fig:overview}.}

\ljc{
We use the same settings as OVD-COCO when conducting experiments on PID except that the pre-trained weight of CLIP visual backbone. As aforementioned in the vanilla method, we use ProductCLIP as the pre-trained weight on PID.
}

\subsection{Comparison with State-of-the-arts.}
\subsubsection{Experiment on OVD-COCO}

\ljc{
We compare all three kinds of basic methods with most existing state-of-the-art methods on OVD-COCO. Table \ref{sota-coco} summarizes the results.} 
\ljc{Although the vanilla method is flexible to replace the detector and classification model, it achieves comparable results on novel categories but obtains bad results on base categories. One possible reason is that CLIP has gaps between images and resized images.}
\xie{Moreover, \xie{CRR} obtains a higher Novel AP$_{50}$ than RegionCLIP, but lower than BARON. However, both RegionCLIP and BARON are with extra backbone, which brings more computational complexity. \xie{DRR} achieves the best results and outperforms BARON by 2.7 Novel AP$_{50}$.}
\ljc{These results demonstrate that \xie{DRR} has the potential to be the best baseline when selecting approaches for open-vocabulary detection. }

\begin{table}[htb]
\caption{\xie{Influence of object localization on OVD-COCO. Faster R-CNN* denotes that it is pre-trained on BigDetection*.}}
      \centering
      \label{det_influence}
      {
        \begin{tabular}{lcccc}
        \toprule
        Method    &  Novel AP$_{50}$ & \lightgray{Base AP$_{50}$} & \lightgray{Overall AP$_{50}$}  \\
        \midrule
        RPN    & 16.6 & \lightgray{20.3} & \lightgray{19.2} \\
        Faster R-CNN    &  19.5 & \lightgray{35.7} & \lightgray{30.9} \\
        OLN    &  26.2 & \lightgray{27.0} & \lightgray{26.6} \\
        Faster R-CNN*    & 29.6 & \lightgray{32.1} & \lightgray{31.1}\\

        \bottomrule
        \end{tabular}
        }
\end{table}

\subsubsection{Experiment on OVD-LVIS}

\ljc{
We further conduct experiments and compare them with state-of-the-art methods on a larger open-vocabulary dataset, \ie OVD-LVIS. We follow the setup of RegionCLIP \cite{regionclip} and provide the results in Table \ref{sota-lvis}. For a fair comparison with the previous SOTA (\ie BARON \cite{BARON}), \xie{we also report the ensemble results which require novel class following BARON. }
\xie{\xie{DRR} achieves 20.1 AP$_r$, which is significantly better than RegionCLIP by 3.0 AP$_r$.}
Meanwhile, \xie{DRR} becomes the new SOTA on OVD-LVIS with the setting of requiring novel class during inference. Similar to OVD-COCO, \xie{CRR} still leads a competitive result.
Instead, the vanilla method obtains bad results compared to other methods, indicating that the operation of crop and resize is non-trivial to recognizing small objects in OVD-LVIS.
}



\subsection{Analysis of Three Fundamental Approaches.}
\ljc{To investigate the behavior of different fundamental approaches of OVD task, we conduct several ablation studies. We design different experiments according to their intrinsic characteristics.}

\begin{table}[htb]
\caption{\xie{Influence of object classification on OVD-COCO.}}
      \centering
      \label{cls_influence}
      {
        \begin{tabular}{lccc}
        \toprule
        Classifier   & Novel AP$_{50}$ & {Base AP$_{50}$} & {Overall AP$_{50}$}  \\
        \midrule
        CLIP-RN50   & 26.6 &\lightgray{27.4} & \lightgray{26.8} \\
        CLIP-ViT-B  & 29.6 & \lightgray{32.1} & \lightgray{31.1} \\
        CLIP-ViT-L  &  31.8 & \lightgray{37.2} & \lightgray{35.6} \\
        \bottomrule
        \end{tabular}
    }
\end{table}

\begin{table}[htb]
\caption{{Effect of image embedding ensemble on OVD-COCO.}}
      \centering
      \label{image_ensemble}
        \resizebox{0.48\textwidth}{!}{
        \begin{tabular}{lccccc}
        \toprule
        Method &  Ensemble & Novel AP$_{50}$ & Base AP$_{50}$ & Overall AP$_{50}$  \\
        \midrule
        Vanilla & $\no$ & 27.3 & \lightgray{28.9} & \lightgray{28.5} \\
        Vanilla & $\yes$ & 29.6 & \lightgray{32.1} & \lightgray{31.1} \\
        \bottomrule
        \end{tabular}
        }
\end{table}

\subsubsection{Vanilla}
\xie{As} this two-stage framework isolates the object localization and classification completely, it can be easily assembled by different models. 
\xie{With the help of advanced object detection models and vision-language models, OVD can be improved flexibly by applying different object localization and object classification modules.}


\begin{table*}[t]
\caption{\ljc{Comparisons of different RPNs over ResNet50 on OVD-COCO. Here, COCO-48-RPN, BD*-489-RPN, and BD*-489-COCO-48-RPN denote that the RPNs are trained on base categories of OVD-COCO, BigDetection, and the combination of base categories of OVD-COCO and BigDetection, respectively.}}
  \label{detector}
  \centering
  \resizebox{1\textwidth}{!}{
    \begin{tabular}{lcccccc}
    \toprule
    Method & Proposal Generation Type & Pre-trained RPN & Multiplying RPN Score & Novel AP$_{50}$ & \lightgray{Base AP$_{50}$} & \lightgray{Overall AP$_{50}$} \\
    \midrule
    \xie{DRR} & RPN & COCO-48-RPN & $\no$ & 31.0 & \lightgray{56.7} & \lightgray{50.0}  \\
    \xie{DRR} & RPN & COCO-48-RPN & $\yes$ & 31.1 & \lightgray{53.9} & \lightgray{48.0}  \\
    \xie{DRR} & Faster R-CNN \small{(class-agnostic)}  & BD*-489-RPN & $\no$ & 30.5 & \lightgray{52.1} & \lightgray{46.5}  \\
    \xie{DRR} & Faster R-CNN \small{(class-agnostic)}  & BD*-489-RPN & $\yes$ & 32.3 & \lightgray{40.4} & \lightgray{38.3}  \\
    \xie{DRR} & Faster R-CNN \small{(class-agnostic)}  & BD*-489-COCO-48-RPN & $\no$  & 30.5  & \lightgray{55.6}  & \lightgray{49.0}  \\
    \xie{DRR} & Faster R-CNN \small{(class-agnostic)}  & BD*-489-COCO-48-RPN & $\yes$  & \textbf{35.8}  & \lightgray{54.6}  & \lightgray{49.4}  \\
    \bottomrule
    \end{tabular}
    }
\end{table*}

\begin{table*}[t]
\caption{\ljc{Effect of CLIP visual backbone on OVD-COCO compared with state-of-the-art methods.}}
  \label{visual_encoder}
  \centering
  \resizebox{0.85\textwidth}{!}{
    \begin{tabular}{ccccccccc}
    \toprule
    Method & Visual Backbone & Detection-tailored Pre-training &
    Novel AP$_{50}$ & \lightgray{Base AP$_{50}$} & \lightgray{Overall AP$_{50}$} \\
    \midrule
    RegionCLIP \cite{regionclip} & ResNet50 & $\no$& 14.2 & \lightgray{52.8} & \lightgray{42.7}  \\
    RegionCLIP \cite{regionclip} & ResNet50 & $\yes$  & 31.4 & \lightgray{57.1} & \lightgray{50.4}  \\
    \xie{DRR} (Ours) & ResNet50 & $\yes$  & \textbf{35.8}  & \lightgray{54.6}  & \lightgray{49.4}  \\
    \midrule
    RegionCLIP \cite{regionclip} & ResNet50x4 & $\yes$  &  39.3 & \lightgray{61.6} & \lightgray{55.7}  \\
    \xie{DRR} (Ours) & ResNet50x4 & $\yes$  & \textbf{41.9} & \lightgray{57.8} & \lightgray{53.7}  \\
    \bottomrule
    \end{tabular}
    }
\end{table*}

\textbf{The influence of object localization.} Table \ref{det_influence} shows the influence of different object localizers. 
\xie{All the experiments use CLIP-ViT-B/32 \cite{CLIP} as the classifier. We attempt several different object localization networks including RPN, Faster R-CNN, and OLN \cite{OLN} with different training schedules. We evaluate the models on the OVD-COCO dataset. As Faster R-CNN can achieve more accurate bounding boxes than RPN, it leads to a higher AP50 on both novel and base categories. Besides, in order to improve the performance for the novel class, we also evaluate the class-agnostic object localization network like OLN. }
OLN learns generalizable objectness and tends to propose any objects in the image.
\xie{Compared with the Faster R-CNN, it outperforms 6.7\% AP50 on novel categories. To obtain better generalization ability, we pre-trained the Faster R-CNN on the BigDetection* dataset. From the fourth row of Table \ref{det_influence}, we observe that the additional pre-training brings a 3.4\% AP50 improvement.}

\xie{
\textbf{The influence of object classification.} Table \ref{cls_influence} shows the influence of different classifiers. According to the above experiments, we use the Faster R-CNN pre-trained on the BigDetection* dataset for object localization. CLIP \cite{CLIP} models are used to classify cropped region proposals. Compared with CLIP-ResNet50 and CLIP-ViT-B/32, CLIP-ViT-L/14 shows better performance. This experiment proves that a strong open-vocabulary object classification model can provide the powerful capability for detecting novel objects.
}


\textbf{The influence of image embedding ensemble.} After localizing objects, we crop and resize the object regions for the \xie{object} classifier to compute image embedding. 
\xie{There are limited contextual cues if cropping directly according to the bounding box localized by the object detector. }
We attempt to expand the bounding boxes by 1.5 times and ensemble the image embeddings from 1× crop and 1.5× crop. 
\xie{We use the Faster R-CNN pre-trained on the BigDetection* dataset for object localization and CLIP-ViT-B/32 for object classification.} 
Table ~\ref{image_ensemble} shows that the image embedding ensemble can improve the OVD performance efficiently.

\subsubsection{\xie{DRR}}




\ljc{
\textbf{The effect of RPN.} As illustrated in Figure \ref{fig:overview}, we first use an offline RPN to obtain proposal bounding boxes and then extract the corresponding features with the help of the visual encoder of CLIP.
In this section, we investigate the effect of different RPN under the settings of \xie{DRR}.
We train several RPNs on different datasets to determine whether a proposal is a foreground. We also replace RPN with Faster R-CNN with a class-agnostic head to improve the detection performance. Compared to RPN, Faster R-CNN prefer to output more accurate bounding boxes with more high-quality objectness logits, thus we attempt to multiply the logits with the final CLIP scores, called ``Multiplying RPN score''. That is, it can be formulated as $\sqrt{s_{1} \cdot s_{2}}$, where $s_{1}$ and $s_{2}$ denote the objectness logits and the CLIP score, respectively.
Table \ref{detector} summarizes the results.
}


\ljc{
From Table \ref{detector}, we get the following observations. 1) Replacing RPN with Faster R-CNN cannot achieve the expected results. Generally, we can improve the overall model performance by improving the offline RPN. However, COCO-48-RPN and BD*-489-COCO-48-RPN obtain similar Novel AP$_{50}$ when without multiplying RPN score (31.0 vs. 30.5), indicating a better offline RPN does not work as well. This is because the offline RPN only provides bounding boxes for the visual backbone of CLIP and does not directly participate in the loss of classification.
2) The significant objectness logits within a better offline RPN are indeed important for model performance. We find that \xie{DRR} with ``multiplying RPN score'' leads to a Novel AP$_{50}$ of 35.8 compared to the one without ``multiplying RPN score'', which becomes a new SOTA over ResNet50 on OVD-COCO.
}

\ljc{
\textbf{The effect of CLIP visual backbones.}
From Table \ref{visual_encoder}, we argue that the detection-tailored pre-training like RegionCLIP \cite{regionclip} is required when needing the best performance.
We then use a larger visual encoder to extract the features of regions of interest. We use ResNet50x4 as the backbone and conduct experiments on OVD-COCO. Relying on the observation from Table \ref{detector}, we use BD*-489-COCO-48-RPN as the RPN and multiply the CLIP scores with RPN scores during inference for \xie{DRR}.
In Table \ref{visual_encoder}, \xie{DRR} surpasses the previous state-of-the-art (\ie RegionCLIP \cite{regionclip}) by 2.6 AP$_{50}$ in novel categories, which benefits from more accurate bounding boxes and more significant objectness logits.
Meanwhile, ResNet50x4 shows better results than ResNet50  from Tables \ref{visual_encoder} \& \ref{sota-coco}. This implies that a model with a larger capacity is able to obtain better representations and improve overall performance. 
}

\begin{figure*}[t]
    \centering
    \includegraphics[width=0.88\textwidth]{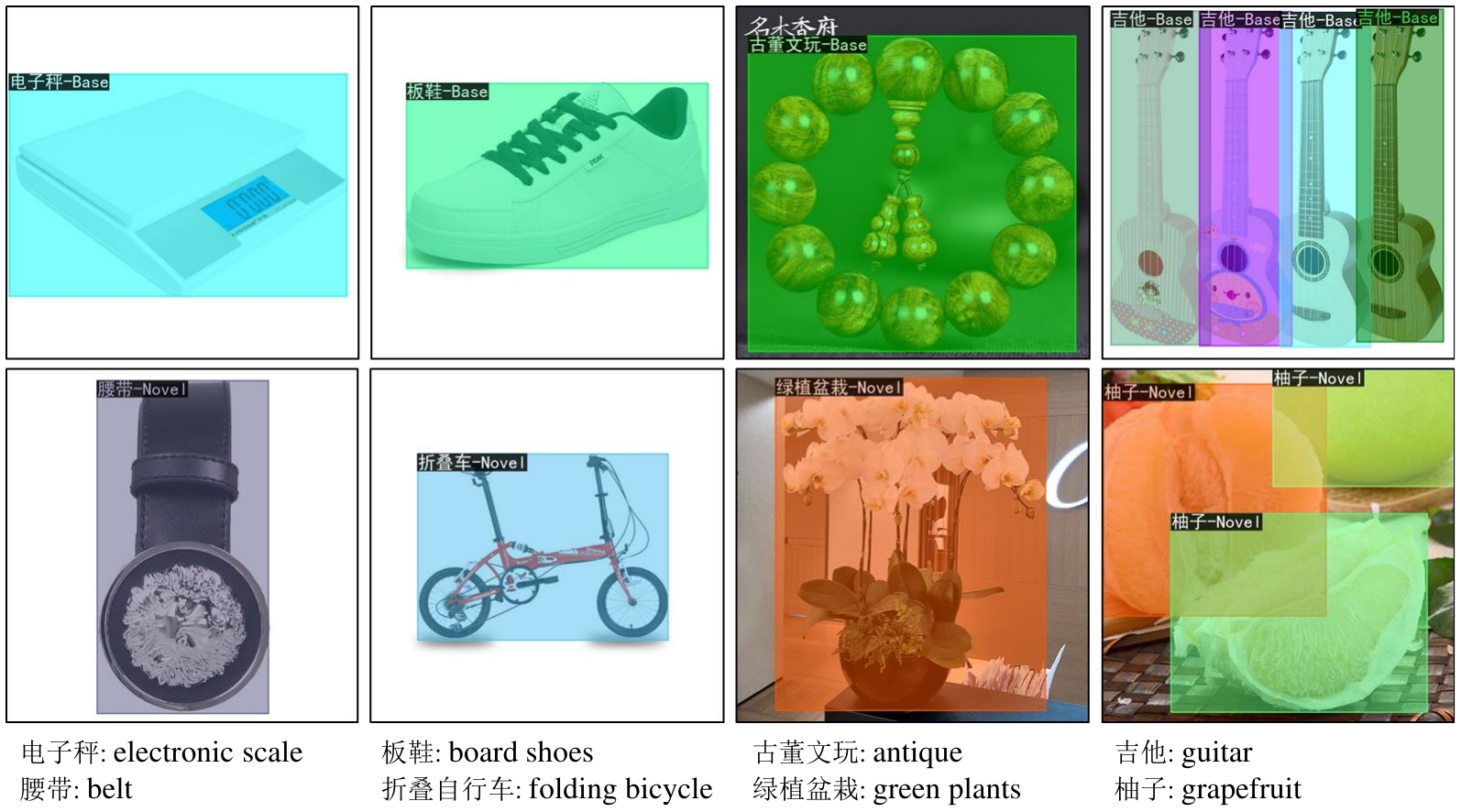} 
    \caption{Examples (with annotations) of PID. The first and second rows are from the base and novel categories, respectively.
    }
    \label{fig:example_PID}
\end{figure*}

\subsubsection{\xie{CRR}}

\textbf{Analysis of Computational efficiency.} Table \ref{fps_cost} shows the performance and computational efficiency of the three frameworks. For a fair comparison, we obtain the results of different models under the same environment. We use the Detectron2 
\footnote{https://github.com/facebookresearch/detectron2}
toolbox based on PyTorch
\footnote{https://pytorch.org}
and \ljc{an} A100 \ljc{GPU} to calculate the FPS. The batch size (denoted as BS in Table) is set to 1. From Table \ref{fps_cost}, we can see that the scheme of sharing the visual backbone has the highest computational efficiency compared to the other two models, and it is far more accurate than the scheme of CLIP on cropped regions. Sharing the visual backbone is indeed more effective in specific real-world scenarios.

\begin{table}[t]
\caption{Comparisons of the computational efficiency over ResNet50 on OVD-COCO.}
      \centering
      \label{fps_cost}
      \resizebox{0.38\textwidth}{!}{
        \begin{tabular}{lcccccc}
        \toprule
        Method &  Params $\downarrow$ & FPS (BS1 A100) $\uparrow$ \\
        \midrule
        Vanilla & 136.9 M  & 2 &  \\
        \xie{DRR} &  143.4 M   &  12 \\
        \xie{CRR} & \textbf{111.6 M} & \textbf{13}\\
        \bottomrule 
        \end{tabular}
        }
\end{table}

\section{Product Image Dataset--A New Object Detection Dataset}
\subsection{Introduction of PID.}\label{PID}
\ljc{
Product Image Dataset (PID) is an object detection dataset consisting of a series of product images along with the corresponding human-annotated bounding boxes and classes.
Note that we label all bounding boxes in Chinese.
PID contains 37,540 images and 52,796 annotations.
The image size of PID is 800 $\times$ 800. 
To adapt to the setting of open vocabulary object detection like OVR-RCNN \cite{ovrcnn}, we split PID into a training dataset and a test dataset, and put more classes in the test dataset. Here, the training dataset consists of 233 classes, 14,802 images, and 20,690 annotations while the test dataset includes 466 classes, 22,738 images, and 32,106 annotations. That is, we use 233 classes as base categories and 233 classes as novel categories. We provide several examples with annotations in Figure \ref{fig:example_PID}. The product image dataset will be released.
}

\begin{table}[t]
\caption{Comparisons of different fundamental approaches over ResNet50 on PID. *More analysis can be found in Section \ref{ft_res_PID}.}
  \label{ft_results_pid}
  \centering
  \resizebox{0.38\textwidth}{!}{
    \begin{tabular}{lcccccccc}
    \toprule
    \multirow{2.5}{*}{Method} &
    \multirow{2.5}{*}{Visual Backbone} &
    \multicolumn{3}{c}{Generalized (233+233)}
    \\ 
    \cmidrule(r){3-5}
    &  & Novel & \lightgray{Base} & \lightgray{Overall} \\
    \midrule
    Vanilla* & ResNet50 & \textbf{42.0} & \lightgray{52.8} & \lightgray{47.4} \\
    \xie{DRR} & ResNet50 & 30.7 & \lightgray{35.6} & \lightgray{33.2}  \\
    \xie{CRR} & ResNet50 & 27.6 & \lightgray{34.3} & \lightgray{31.0} \\
    \bottomrule
    \end{tabular}
    }
\end{table}

\subsection{Finetuned results on PID.}
\label{ft_res_PID}
\ljc{
We train three approaches on the base categories of PID and then evaluate them on the base and novel categories. We adopt ResNet50 as the visual backbone of CLIP.
The results are shown in Table \ref{ft_results_pid}.
Somewhat surprisingly, the vanilla method achieves the best performance compared to \xie{DRR} and \xie{CRR}, which is inconsistent with the results on OVD-COCO and OVD-LVIS. One possible reason is that the objects in PID are large such that cropping and resizing the proposal is an effective way to improve the overall performance.
}

\begin{table*}[t]
\caption{\ljc{Effect of RPNs and the ensemble score on PID.}}
  \label{multiply_rpn_score_PID}
  \centering
  \resizebox{0.7\textwidth}{!}{
    \begin{tabular}{ccccccccc}
    \toprule
    \multirow{2.5}{*}{Visual Encoder Pre-training} &
    \multirow{2.5}{*}{Pre-trained RPN} &
    \multirow{2.5}{*}{Multiplying RPN Score} &
    \multicolumn{3}{c}{Generalized (233+233)}
    \\ 
    \cmidrule(r){4-6}
    &  & & Novel & \lightgray{Base} & \lightgray{Overall}   \\
    \midrule
    ProductCLIP & PID-233-RPN & $\no$ & 22.0 & \lightgray{34.1} & \lightgray{28.1} \\
    ProductCLIP & PID-233-RPN & $\yes$ & 26.6 & \lightgray{36.2} & \lightgray{31.4} \\
    ProductCLIP & BD*-489-PID-233-RPN & $\yes$ & 30.7 & \lightgray{35.6} & \lightgray{33.2} \\
    \bottomrule
    \end{tabular}
    }
\end{table*}

\begin{table*}[t]
\caption{\ljc{Effect of knowledge distillation on PID. We use L1 distillation by default.}}
  \label{KD_PID}
  \centering
  \resizebox{0.7\textwidth}{!}{
    \begin{tabular}{ccccccccc}
    \toprule
    \multirow{2.5}{*}{Visual Encoder Pre-training} &
    \multirow{2.5}{*}{Pre-trained RPN} &
    \multirow{2.5}{*}{Temperature} &
    \multicolumn{3}{c}{Generalized (233+233)}
    \\ 
    \cmidrule(r){4-6}
    &  & & Novel & \lightgray{Base} & \lightgray{Overall}  \\
    \midrule
    ProductCLIP & PID-233-RPN & - & 22.0 & \lightgray{34.1} & \lightgray{28.1} \\
    ProductCLIP & PID-233-RPN & 1 & 22.6 & \lightgray{34.5} & \lightgray{28.6} \\
    ProductCLIP & PID-233-RPN & 10 & 21.1 & \lightgray{31.7} & \lightgray{26.4} \\
    ProductCLIP & PID-233-RPN & 100 & 16.3 & \lightgray{24.5} & \lightgray{20.4} \\
    
    \bottomrule
    \end{tabular}
    }
\end{table*}

\begin{table*}[t]
\caption{\xie{Effect of the number of prompts of the text encoder on PID.}}
  \label{num_prompt_PID}
  \centering
  \resizebox{0.7\textwidth}{!}{
    \begin{tabular}{ccccccccc}
    \toprule
    \multirow{2.5}{*}{Visual Encoder Pre-training} &
    \multirow{2.5}{*}{Pre-trained RPN} &
    \multirow{2.5}{*}{\#prompt} &
    \multicolumn{3}{c}{Generalized (233+233)}
    \\ 
    \cmidrule(r){4-6}
    &  & & Novel & \lightgray{Base} & \lightgray{Overall}  \\
    \midrule
    ProductCLIP & PID-233-RPN & 93 & 22.0 & \lightgray{34.1} & \lightgray{28.1} \\
    ProductCLIP & PID-233-RPN & 1 & 22.3 & \lightgray{34.3} & \lightgray{28.3} \\
    \bottomrule
    \end{tabular}
    }
\end{table*}

\begin{table*}[t]
\caption{\ljc{Effect of prompt tuning on PID.}}
  \label{prompt_tuning_PID}
  \centering
  \resizebox{0.8\textwidth}{!}{
    \begin{tabular}{ccccccccc}
    \toprule
    \multirow{2.5}{*}{Visual Encoder Pre-training} &
    \multirow{2.5}{*}{Pre-trained RPN} &
    \multirow{2.5}{*}{Prompt Tuning} &
    \multicolumn{3}{c}{Generalized (233+233)}
    \\ 
    \cmidrule(r){4-6}
    &  & & Novel & \lightgray{Base} & \lightgray{Overall}  \\
    \midrule
    ProductCLIP & BD*-489-PID-233-RPN  & $\no$ & 27.1 & \lightgray{35.8} & \lightgray{31.5} \\
    ProductCLIP & BD*-489-PID-233-RPN  & $\yes$ & 28.1 & \lightgray{35.6} & \lightgray{31.9} \\
    \bottomrule
    \end{tabular}
    }
\end{table*}

\subsection{Ablation study on PID.}\label{PID_Ablation}
\ljc{
\textbf{Summary of ablation study.} Several related works \cite{ViLD, regionclip, bangalath2022bridging, BARON} in open-vocabulary detection have applied a variety of techniques to improve the model performance. However, it is worth noting that some works use a part of these techniques while others do not, which makes this line of work confusing. That is, it is hard to use existing techniques to improve the model performance when transferring to a new dataset, \eg PID. Here, we investigate the effectiveness of commonly used techniques on PID, including different proposal generators, multiplying RPN score, knowledge distillation, and prompt learning. We use DRR to conduct all experiments on PID.
}

\ljc{
\textbf{Why use \xie{DRR} to conduct ablation study on PID?} From Table \ref{ft_results_pid}, it seems that the vanilla method is a better choice compared to \xie{CRR} and \xie{DRR}. However, it suffers from slow inference speed \cite{ViLD} and achieves unsatisfactory results on the two popular public OVD datasets, \ie OVD-COCO and OVD-LVIS (see Tables \ref{sota-coco} \& \ref{sota-lvis} \& \ref{ft_results_pid}). Instead, \xie{DRR} outperforms the vanilla method and \xie{CRR} on OVD-COCO and OVD-LVIS and is with excellent computational efficiency (see Table \ref{fps_cost}). Since the main goal is to investigate which technique is more effective when transferring to new datasets, we adopt DRR to conduct the ablation study on PID.
}

\ljc{
\textbf{The effect of RPNs and the ensemble score.} 
\xie{First, we train a CLIP model using image-text product pairs and name it ProductCLIP.} Note that the text encoder of ProductCLIP is designed for Chinese. We then train an offline RPN on the base categories of PID for \xie{DRR}, called PID-233-RPN. Similar to Table \ref{detector}, we investigate the effectiveness of ``multiplying RPN score'' during inference. We also train a BD*-489-PID-233-RPN to further improve the quality of bounding boxes and objectness logits. From Table \ref{multiply_rpn_score_PID}, the model with ``multiplying RPN score'' can achieve better detection performance over base and novel categories. When improving the performance of the offline RPN, the Novel AP$_{50}$ improves from 26.6 to 30.7. We can obtain similar observations as in Table \ref{detector}.
}

\ljc{
\textbf{The effect of knowledge distillation.}
Knowledge distillation is a commonly used technique to bridge the gap between the object-level representations from detectors and the image-level representations from CLIP.
Thus, they can obtain significant classification scores from the object-level representations and the embeddings from the text encoder of CLIP.
Following \cite{ViLD,bangalath2022bridging}, we use L1 knowledge distillation to check whether this technique is valid for PID.
We also adjust different values of temperature in knowledge distillation to find a good hyperparameter.
Table \ref{KD_PID} summarizes the results.
We observe that \xie{DRR} with knowledge distillation (KD) achieves slightly better results than the one without KD, \ie 22.6 vs. 22.0. One possible reason is that we pre-train the visual backbone of CLIP using image-text product pairs, that is, there is no gap between the object-level representations from the visual backbone with the text encoder.
}

\ljc{
\xie{\textbf{The effect of prompt numbers.} }
In the setup of open-vocabulary detection, we usually obtain the text embeddings by directly feeding the concepts with human-crafted prompts into the text encoder of CLIP. 
\xie{Here, We evaluate different number of prompts of the text encoder in Table \ref{num_prompt_PID}.}
We observe that the effectiveness of the number of prompts can be negligible on PID.
}

\ljc{
\textbf{The effect of prompt tuning.} Next, we introduce another common technique in prompt modification. We replace the human-crafted prompt with a learned prompt by prompt tuning \cite{zhou2022learning}. Table \ref{prompt_tuning_PID} summarizes the results. The models with prompt tuning can improve by 1.0 Novel AP$_{50}$, indicating that prompt tuning is an effective technique to improve the model performance.
}






\section{Conclusion and Social Impacts}
\label{conclusion}
In this paper, we have conducted a comprehensive study of three fundamental commonly used approaches for open-vocabulary detection, a vanilla method, decoupling RPN and ROI head (\xie{DRR}), and coupling RPN and ROI head (\xie{CRR}). We analyze the advantages and the disadvantages of the three approaches and give a discussion about how and when to use them. Extensive experiments demonstrate the effectiveness of different approaches under different setups on COCO and LVIS benchmarks. Besides, we propose a product dataset for object detection called PID and provide a strong baseline on PID. We hope our analyses and datasets promote the development of open-vocabulary detection.

\textbf{Acknowledgement.} This work was supported in part by the National Key Research and Development Program of China under Grant 2018AAA0100405. We thank Bang Yang for the help in collecting CC3M. We also thank DeXin Wang and Dan Zhou for the help in collecting PID.

\bibliographystyle{ACM-Reference-Format}
\bibliography{references}



\end{document}

%% file: math_commands.tex
\usepackage{CJKutf8}

\usepackage[utf8]{inputenc} 
\usepackage[T1]{fontenc}    
\usepackage{url}            
\usepackage{nicefrac}       
\usepackage{lipsum}		
\usepackage{doi}

\usepackage{xcolor}         

\usepackage{times}
\usepackage{microtype} %
\usepackage{balance} 

\usepackage{helvet}
\usepackage{courier}

\usepackage{makecell}
\usepackage{graphicx}
\usepackage{color}
\usepackage{amsfonts}
\usepackage{amsmath}

\usepackage{multirow}
\usepackage{booktabs}
\usepackage{wrapfig}

\def\ie{\mbox{\textit{i.e.}, }}
\def\eg{\mbox{\textit{e.g.}, }}

\DeclareMathAlphabet\mathbfcal{OMS}{cmsy}{b}{n}

\def\0{{\bf 0}}
\usepackage{ dsfont }
\def\1{\mathds{1}}

\newtheorem*{*thm}{Theorem}

\newtheorem*{*lemma}{Lemma}

\usepackage{enumitem}

\newcommand{\yes}{\textcolor[rgb]{0,0.6,0}{\large{\checkmark}}}
\newcommand{\no}{\textcolor[rgb]{0.6,0,0}{\large{{\bf\times}}}}

\usepackage{array}
\usepackage{tabu}
\makeatletter
\newcommand{\nipstophline}{%
	\noalign {\ifnum 0=`}\fi \hrule height 4pt
	\futurelet \reserved@a \@xhline
}
\newcommand{\nipsbottomhline}{%
	\noalign {\ifnum 0=`}\fi \hrule height 1pt
	\futurelet \reserved@a \@xhline
}

\def\ljc{\textcolor{black}}
\def\xie{\textcolor{black}}
\def\wxy{\textcolor{black}}

\def\lightgray{\textcolor{gray}}

%% file: sample-sigconf.bbl

\begin{thebibliography}{28}


\ifx \showCODEN    \undefined \def \showCODEN     #1{\unskip}     \fi
\ifx \showDOI      \undefined \def \showDOI       #1{#1}\fi
\ifx \showISBNx    \undefined \def \showISBNx     #1{\unskip}     \fi
\ifx \showISBNxiii \undefined \def \showISBNxiii  #1{\unskip}     \fi
\ifx \showISSN     \undefined \def \showISSN      #1{\unskip}     \fi
\ifx \showLCCN     \undefined \def \showLCCN      #1{\unskip}     \fi
\ifx \shownote     \undefined \def \shownote      #1{#1}          \fi
\ifx \showarticletitle \undefined \def \showarticletitle #1{#1}   \fi
\ifx \showURL      \undefined \def \showURL       {\relax}        \fi
\providecommand\bibfield[2]{#2}
\providecommand\bibinfo[2]{#2}
\providecommand\natexlab[1]{#1}
\providecommand\showeprint[2][]{arXiv:#2}

\bibitem[Bangalath et~al\mbox{.}(2022)]%
        {bangalath2022bridging}
\bibfield{author}{\bibinfo{person}{Hanoona Bangalath},
  \bibinfo{person}{Muhammad Maaz}, \bibinfo{person}{Muhammad~Uzair Khattak},
  \bibinfo{person}{Salman~H Khan}, {and} \bibinfo{person}{Fahad Shahbaz~Khan}.}
  \bibinfo{year}{2022}\natexlab{}.
\newblock \showarticletitle{Bridging the gap between object and image-level
  representations for open-vocabulary detection}.
\newblock \bibinfo{journal}{\emph{Advances in Neural Information Processing
  Systems}}  \bibinfo{volume}{35} (\bibinfo{year}{2022}).
\newblock


\bibitem[Cai et~al\mbox{.}(2022)]%
        {bigdetection}
\bibfield{author}{\bibinfo{person}{Likun Cai}, \bibinfo{person}{Zhi Zhang},
  \bibinfo{person}{Yi Zhu}, \bibinfo{person}{Li Zhang}, \bibinfo{person}{Mu
  Li}, {and} \bibinfo{person}{Xiangyang Xue}.} \bibinfo{year}{2022}\natexlab{}.
\newblock \showarticletitle{BigDetection: A Large-scale Benchmark for Improved
  Object Detector Pre-training}. In \bibinfo{booktitle}{\emph{Proceedings of
  the IEEE/CVF Conference on Computer Vision and Pattern Recognition}}.
  \bibinfo{pages}{4777--4787}.
\newblock


\bibitem[Chen et~al\mbox{.}(2022)]%
        {chen2022open}
\bibfield{author}{\bibinfo{person}{Peixian Chen}, \bibinfo{person}{Kekai
  Sheng}, \bibinfo{person}{Mengdan Zhang}, \bibinfo{person}{Yunhang Shen},
  \bibinfo{person}{Ke Li}, {and} \bibinfo{person}{Chunhua Shen}.}
  \bibinfo{year}{2022}\natexlab{}.
\newblock \showarticletitle{Open Vocabulary Object Detection with Proposal
  Mining and Prediction Equalization}.
\newblock \bibinfo{journal}{\emph{arXiv preprint arXiv:2206.11134}}
  (\bibinfo{year}{2022}).
\newblock


\bibitem[Du et~al\mbox{.}(2022)]%
        {du2022learning}
\bibfield{author}{\bibinfo{person}{Yu Du}, \bibinfo{person}{Fangyun Wei},
  \bibinfo{person}{Zihe Zhang}, \bibinfo{person}{Miaojing Shi},
  \bibinfo{person}{Yue Gao}, {and} \bibinfo{person}{Guoqi Li}.}
  \bibinfo{year}{2022}\natexlab{}.
\newblock \showarticletitle{Learning to prompt for open-vocabulary object
  detection with vision-language model}. In
  \bibinfo{booktitle}{\emph{Proceedings of the IEEE/CVF Conference on Computer
  Vision and Pattern Recognition}}. \bibinfo{pages}{14084--14093}.
\newblock


\bibitem[Feng et~al\mbox{.}(2022)]%
        {feng2022promptdet}
\bibfield{author}{\bibinfo{person}{Chengjian Feng}, \bibinfo{person}{Yujie
  Zhong}, \bibinfo{person}{Zequn Jie}, \bibinfo{person}{Xiangxiang Chu},
  \bibinfo{person}{Haibing Ren}, \bibinfo{person}{Xiaolin Wei},
  \bibinfo{person}{Weidi Xie}, {and} \bibinfo{person}{Lin Ma}.}
  \bibinfo{year}{2022}\natexlab{}.
\newblock \showarticletitle{Promptdet: Expand your detector vocabulary with
  uncurated images}. In \bibinfo{booktitle}{\emph{European Conference on
  Computer Vision}}.
\newblock


\bibitem[Gao et~al\mbox{.}(2022)]%
        {gao2022open}
\bibfield{author}{\bibinfo{person}{Mingfei Gao}, \bibinfo{person}{Chen Xing},
  \bibinfo{person}{Juan~Carlos Niebles}, \bibinfo{person}{Junnan Li},
  \bibinfo{person}{Ran Xu}, \bibinfo{person}{Wenhao Liu}, {and}
  \bibinfo{person}{Caiming Xiong}.} \bibinfo{year}{2022}\natexlab{}.
\newblock \showarticletitle{Open vocabulary object detection with pseudo
  bounding-box labels}. In \bibinfo{booktitle}{\emph{European Conference on
  Computer Vision}}. \bibinfo{pages}{266--282}.
\newblock


\bibitem[Gu et~al\mbox{.}(2022)]%
        {ViLD}
\bibfield{author}{\bibinfo{person}{Xiuye Gu}, \bibinfo{person}{Tsung-Yi Lin},
  \bibinfo{person}{Weicheng Kuo}, {and} \bibinfo{person}{Yin Cui}.}
  \bibinfo{year}{2022}\natexlab{}.
\newblock \showarticletitle{Open-vocabulary object detection via vision and
  language knowledge distillation}. In \bibinfo{booktitle}{\emph{International
  Conference on Learning Representations}}.
\newblock


\bibitem[Gupta et~al\mbox{.}(2019)]%
        {LVIS}
\bibfield{author}{\bibinfo{person}{Agrim Gupta}, \bibinfo{person}{Piotr
  Dollar}, {and} \bibinfo{person}{Ross Girshick}.}
  \bibinfo{year}{2019}\natexlab{}.
\newblock \showarticletitle{Lvis: A dataset for large vocabulary instance
  segmentation}. In \bibinfo{booktitle}{\emph{Proceedings of the IEEE/CVF
  Conference on Computer Vision and Pattern Recognition}}.
  \bibinfo{pages}{5356--5364}.
\newblock


\bibitem[Jia et~al\mbox{.}(2021)]%
        {jia2021scaling}
\bibfield{author}{\bibinfo{person}{Chao Jia}, \bibinfo{person}{Yinfei Yang},
  \bibinfo{person}{Ye Xia}, \bibinfo{person}{Yi-Ting Chen},
  \bibinfo{person}{Zarana Parekh}, \bibinfo{person}{Hieu Pham},
  \bibinfo{person}{Quoc Le}, \bibinfo{person}{Yun-Hsuan Sung},
  \bibinfo{person}{Zhen Li}, {and} \bibinfo{person}{Tom Duerig}.}
  \bibinfo{year}{2021}\natexlab{}.
\newblock \showarticletitle{Scaling up visual and vision-language
  representation learning with noisy text supervision}. In
  \bibinfo{booktitle}{\emph{International Conference on Machine Learning}}.
  \bibinfo{pages}{4904--4916}.
\newblock


\bibitem[Kim et~al\mbox{.}(2022)]%
        {OLN}
\bibfield{author}{\bibinfo{person}{Dahun Kim}, \bibinfo{person}{Tsung-Yi Lin},
  \bibinfo{person}{Anelia Angelova}, \bibinfo{person}{In~So Kweon}, {and}
  \bibinfo{person}{Weicheng Kuo}.} \bibinfo{year}{2022}\natexlab{}.
\newblock \showarticletitle{Learning open-world object proposals without
  learning to classify}.
\newblock \bibinfo{journal}{\emph{IEEE Robotics and Automation Letters}}
  \bibinfo{volume}{7}, \bibinfo{number}{2} (\bibinfo{year}{2022}),
  \bibinfo{pages}{5453--5460}.
\newblock


\bibitem[Kuo et~al\mbox{.}(2023)]%
        {kuo2022f}
\bibfield{author}{\bibinfo{person}{Weicheng Kuo}, \bibinfo{person}{Yin Cui},
  \bibinfo{person}{Xiuye Gu}, \bibinfo{person}{AJ Piergiovanni}, {and}
  \bibinfo{person}{Anelia Angelova}.} \bibinfo{year}{2023}\natexlab{}.
\newblock \showarticletitle{F-VLM: Open-Vocabulary Object Detection upon Frozen
  Vision and Language Models}. In \bibinfo{booktitle}{\emph{International
  Conference on Learning Representations}}.
\newblock


\bibitem[Lin et~al\mbox{.}(2014)]%
        {COCO}
\bibfield{author}{\bibinfo{person}{Tsung-Yi Lin}, \bibinfo{person}{Michael
  Maire}, \bibinfo{person}{Serge Belongie}, \bibinfo{person}{James Hays},
  \bibinfo{person}{Pietro Perona}, \bibinfo{person}{Deva Ramanan},
  \bibinfo{person}{Piotr Doll{\'a}r}, {and} \bibinfo{person}{C~Lawrence
  Zitnick}.} \bibinfo{year}{2014}\natexlab{}.
\newblock \showarticletitle{Microsoft coco: Common objects in context}. In
  \bibinfo{booktitle}{\emph{European Conference on Computer Vision}}.
  \bibinfo{pages}{740--755}.
\newblock


\bibitem[Ma et~al\mbox{.}(2022)]%
        {ma2022open}
\bibfield{author}{\bibinfo{person}{Zongyang Ma}, \bibinfo{person}{Guan Luo},
  \bibinfo{person}{Jin Gao}, \bibinfo{person}{Liang Li}, \bibinfo{person}{Yuxin
  Chen}, \bibinfo{person}{Shaoru Wang}, \bibinfo{person}{Congxuan Zhang}, {and}
  \bibinfo{person}{Weiming Hu}.} \bibinfo{year}{2022}\natexlab{}.
\newblock \showarticletitle{Open-vocabulary one-stage detection with
  hierarchical visual-language knowledge distillation}. In
  \bibinfo{booktitle}{\emph{Proceedings of the IEEE/CVF Conference on Computer
  Vision and Pattern Recognition}}. \bibinfo{pages}{14074--14083}.
\newblock


\bibitem[Radford et~al\mbox{.}(2021)]%
        {CLIP}
\bibfield{author}{\bibinfo{person}{Alec Radford}, \bibinfo{person}{Jong~Wook
  Kim}, \bibinfo{person}{Chris Hallacy}, \bibinfo{person}{Aditya Ramesh},
  \bibinfo{person}{Gabriel Goh}, \bibinfo{person}{Sandhini Agarwal},
  \bibinfo{person}{Girish Sastry}, \bibinfo{person}{Amanda Askell},
  \bibinfo{person}{Pamela Mishkin}, \bibinfo{person}{Jack Clark},
  {et~al\mbox{.}}} \bibinfo{year}{2021}\natexlab{}.
\newblock \showarticletitle{Learning transferable visual models from natural
  language supervision}. In \bibinfo{booktitle}{\emph{International Conference
  on Machine Learning}}. \bibinfo{pages}{8748--8763}.
\newblock


\bibitem[Redmon et~al\mbox{.}(2016)]%
        {redmon2016you}
\bibfield{author}{\bibinfo{person}{Joseph Redmon}, \bibinfo{person}{Santosh
  Divvala}, \bibinfo{person}{Ross Girshick}, {and} \bibinfo{person}{Ali
  Farhadi}.} \bibinfo{year}{2016}\natexlab{}.
\newblock \showarticletitle{You only look once: Unified, real-time object
  detection}. In \bibinfo{booktitle}{\emph{Proceedings of the IEEE/CVF
  Conference on Computer Vision and Pattern Recognition}}.
  \bibinfo{pages}{779--788}.
\newblock


\bibitem[Ren et~al\mbox{.}(2015)]%
        {fasterrcnn}
\bibfield{author}{\bibinfo{person}{Shaoqing Ren}, \bibinfo{person}{Kaiming He},
  \bibinfo{person}{Ross Girshick}, {and} \bibinfo{person}{Jian Sun}.}
  \bibinfo{year}{2015}\natexlab{}.
\newblock \showarticletitle{Faster r-cnn: Towards real-time object detection
  with region proposal networks}.
\newblock \bibinfo{journal}{\emph{Advances in Neural Information Processing
  Systems}}  \bibinfo{volume}{28} (\bibinfo{year}{2015}).
\newblock


\bibitem[Russakovsky et~al\mbox{.}(2015)]%
        {imagenet}
\bibfield{author}{\bibinfo{person}{Olga Russakovsky}, \bibinfo{person}{Jia
  Deng}, \bibinfo{person}{Hao Su}, \bibinfo{person}{Jonathan Krause},
  \bibinfo{person}{Sanjeev Satheesh}, \bibinfo{person}{Sean Ma},
  \bibinfo{person}{Zhiheng Huang}, \bibinfo{person}{Andrej Karpathy},
  \bibinfo{person}{Aditya Khosla}, \bibinfo{person}{Michael Bernstein},
  {et~al\mbox{.}}} \bibinfo{year}{2015}\natexlab{}.
\newblock \showarticletitle{Imagenet large scale visual recognition challenge}.
\newblock \bibinfo{journal}{\emph{International Journal of Computer Vision}}
  \bibinfo{volume}{115} (\bibinfo{year}{2015}), \bibinfo{pages}{211--252}.
\newblock


\bibitem[Sharma et~al\mbox{.}(2018)]%
        {CC3M}
\bibfield{author}{\bibinfo{person}{Piyush Sharma}, \bibinfo{person}{Nan Ding},
  \bibinfo{person}{Sebastian Goodman}, {and} \bibinfo{person}{Radu Soricut}.}
  \bibinfo{year}{2018}\natexlab{}.
\newblock \showarticletitle{Conceptual captions: A cleaned, hypernymed, image
  alt-text dataset for automatic image captioning}. In
  \bibinfo{booktitle}{\emph{Proceedings of the 56th Annual Meeting of the
  Association for Computational Linguistics (Volume 1: Long Papers)}}.
  \bibinfo{pages}{2556--2565}.
\newblock


\bibitem[Wang et~al\mbox{.}(2023)]%
        {wang2023detecting}
\bibfield{author}{\bibinfo{person}{Zhenyu Wang}, \bibinfo{person}{Yali Li},
  \bibinfo{person}{Xi Chen}, \bibinfo{person}{Ser-Nam Lim},
  \bibinfo{person}{Antonio Torralba}, \bibinfo{person}{Hengshuang Zhao}, {and}
  \bibinfo{person}{Shengjin Wang}.} \bibinfo{year}{2023}\natexlab{}.
\newblock \showarticletitle{Detecting Everything in the Open World: Towards
  Universal Object Detection}. In \bibinfo{booktitle}{\emph{Proceedings of the
  IEEE/CVF Conference on Computer Vision and Pattern Recognition}}.
\newblock


\bibitem[Wu et~al\mbox{.}(2023a)]%
        {BARON}
\bibfield{author}{\bibinfo{person}{Size Wu}, \bibinfo{person}{Wenwei Zhang},
  \bibinfo{person}{Sheng Jin}, \bibinfo{person}{Wentao Liu}, {and}
  \bibinfo{person}{Chen~Change Loy}.} \bibinfo{year}{2023}\natexlab{a}.
\newblock \showarticletitle{Aligning bag of regions for open-vocabulary object
  detection}. In \bibinfo{booktitle}{\emph{Proceedings of the IEEE/CVF
  Conference on Computer Vision and Pattern Recognition}}.
\newblock


\bibitem[Wu et~al\mbox{.}(2023b)]%
        {wu2023cora}
\bibfield{author}{\bibinfo{person}{Xiaoshi Wu}, \bibinfo{person}{Feng Zhu},
  \bibinfo{person}{Rui Zhao}, {and} \bibinfo{person}{Hongsheng Li}.}
  \bibinfo{year}{2023}\natexlab{b}.
\newblock \showarticletitle{CORA: Adapting CLIP for Open-Vocabulary Detection
  with Region Prompting and Anchor Pre-Matching}. In
  \bibinfo{booktitle}{\emph{Proceedings of the IEEE/CVF Conference on Computer
  Vision and Pattern Recognition}}.
\newblock


\bibitem[Xie et~al\mbox{.}(2022)]%
        {xie2022zero}
\bibfield{author}{\bibinfo{person}{Chunyu Xie}, \bibinfo{person}{Heng Cai},
  \bibinfo{person}{Jianfei Song}, \bibinfo{person}{Jincheng Li},
  \bibinfo{person}{Fanjing Kong}, \bibinfo{person}{Xiaoyu Wu},
  \bibinfo{person}{Henrique Morimitsu}, \bibinfo{person}{Lin Yao},
  \bibinfo{person}{Dexin Wang}, \bibinfo{person}{Dawei Leng}, {et~al\mbox{.}}}
  \bibinfo{year}{2022}\natexlab{}.
\newblock \showarticletitle{Zero and R2D2: A Large-scale Chinese Cross-modal
  Benchmark and A Vision-Language Framework}.
\newblock \bibinfo{journal}{\emph{arXiv preprint arXiv:2205.03860}}
  (\bibinfo{year}{2022}).
\newblock


\bibitem[Yang et~al\mbox{.}(2022)]%
        {yang2022chinese}
\bibfield{author}{\bibinfo{person}{An Yang}, \bibinfo{person}{Junshu Pan},
  \bibinfo{person}{Junyang Lin}, \bibinfo{person}{Rui Men},
  \bibinfo{person}{Yichang Zhang}, \bibinfo{person}{Jingren Zhou}, {and}
  \bibinfo{person}{Chang Zhou}.} \bibinfo{year}{2022}\natexlab{}.
\newblock \showarticletitle{Chinese CLIP: Contrastive Vision-Language
  Pretraining in Chinese}.
\newblock \bibinfo{journal}{\emph{arXiv preprint arXiv:2211.01335}}
  (\bibinfo{year}{2022}).
\newblock


\bibitem[Zareian et~al\mbox{.}(2021)]%
        {ovrcnn}
\bibfield{author}{\bibinfo{person}{Alireza Zareian},
  \bibinfo{person}{Kevin~Dela Rosa}, \bibinfo{person}{Derek~Hao Hu}, {and}
  \bibinfo{person}{Shih-Fu Chang}.} \bibinfo{year}{2021}\natexlab{}.
\newblock \showarticletitle{Open-vocabulary object detection using captions}.
  In \bibinfo{booktitle}{\emph{Proceedings of the IEEE/CVF Conference on
  Computer Vision and Pattern Recognition}}. \bibinfo{pages}{14393--14402}.
\newblock


\bibitem[Zhao et~al\mbox{.}(2022)]%
        {zhao2022exploiting}
\bibfield{author}{\bibinfo{person}{Shiyu Zhao}, \bibinfo{person}{Zhixing
  Zhang}, \bibinfo{person}{Samuel Schulter}, \bibinfo{person}{Long Zhao},
  \bibinfo{person}{BG Vijay~Kumar}, \bibinfo{person}{Anastasis Stathopoulos},
  \bibinfo{person}{Manmohan Chandraker}, {and} \bibinfo{person}{Dimitris~N
  Metaxas}.} \bibinfo{year}{2022}\natexlab{}.
\newblock \showarticletitle{Exploiting unlabeled data with vision and language
  models for object detection}. In \bibinfo{booktitle}{\emph{European
  Conference on Computer Vision}}. \bibinfo{pages}{159--175}.
\newblock


\bibitem[Zhong et~al\mbox{.}(2022)]%
        {regionclip}
\bibfield{author}{\bibinfo{person}{Yiwu Zhong}, \bibinfo{person}{Jianwei Yang},
  \bibinfo{person}{Pengchuan Zhang}, \bibinfo{person}{Chunyuan Li},
  \bibinfo{person}{Noel Codella}, \bibinfo{person}{Liunian~Harold Li},
  \bibinfo{person}{Luowei Zhou}, \bibinfo{person}{Xiyang Dai},
  \bibinfo{person}{Lu Yuan}, \bibinfo{person}{Yin Li}, {et~al\mbox{.}}}
  \bibinfo{year}{2022}\natexlab{}.
\newblock \showarticletitle{Regionclip: Region-based language-image
  pretraining}. In \bibinfo{booktitle}{\emph{Proceedings of the IEEE/CVF
  Conference on Computer Vision and Pattern Recognition}}.
  \bibinfo{pages}{16793--16803}.
\newblock


\bibitem[Zhou et~al\mbox{.}(2022b)]%
        {zhou2022learning}
\bibfield{author}{\bibinfo{person}{Kaiyang Zhou}, \bibinfo{person}{Jingkang
  Yang}, \bibinfo{person}{Chen~Change Loy}, {and} \bibinfo{person}{Ziwei Liu}.}
  \bibinfo{year}{2022}\natexlab{b}.
\newblock \showarticletitle{Learning to prompt for vision-language models}.
\newblock \bibinfo{journal}{\emph{International Journal of Computer Vision}}
  \bibinfo{volume}{130}, \bibinfo{number}{9} (\bibinfo{year}{2022}),
  \bibinfo{pages}{2337--2348}.
\newblock


\bibitem[Zhou et~al\mbox{.}(2022a)]%
        {Detic}
\bibfield{author}{\bibinfo{person}{Xingyi Zhou}, \bibinfo{person}{Rohit
  Girdhar}, \bibinfo{person}{Armand Joulin}, \bibinfo{person}{Philipp
  Kr{\"a}henb{\"u}hl}, {and} \bibinfo{person}{Ishan Misra}.}
  \bibinfo{year}{2022}\natexlab{a}.
\newblock \showarticletitle{Detecting twenty-thousand classes using image-level
  supervision}. In \bibinfo{booktitle}{\emph{European Conference on Computer
  Vision}}. \bibinfo{pages}{350--368}.
\newblock


\end{thebibliography}
